\documentclass[runningheads]{llncs}
\usepackage{graphicx}
\usepackage{amsmath}
\usepackage{makecell}
\usepackage{bm}
\usepackage{xcolor}

\begin{document}
\title{From Connectomic to Task-evoked Fingerprints: Individualized Prediction of Task Contrasts from Resting-state Functional Connectivity}
\titlerunning{From Connectomic to Task-evoked Fingerprints}

\author{Gia H. Ngo\inst{1} \and
Meenakshi Khosla\inst{1} \and
Keith Jamison\inst{2} \and
Amy Kuceyeski\inst{2,3} \and
Mert R. Sabuncu\inst{1,2,4}}
\authorrunning{Ngo et al.}

\institute{Princeton University, Princeton NJ 08544, USA \and
Springer Heidelberg, Tiergartenstr. 17, 69121 Heidelberg, Germany
\email{lncs@springer.com}\\
\url{http://www.springer.com/gp/computer-science/lncs} \and
ABC Institute, Rupert-Karls-University Heidelberg, Heidelberg, Germany\\
\email{\{abc,lncs\}@uni-heidelberg.de}}

\institute{School of Electrical \& Computer Engineering, Cornell University \and
Radiology, Weill Cornell Medicine \and
Brain and Mind Research Institute, Weill Cornell Medicine \and
Nancy E. \& Peter C. Meinig School of Biomedical Engineering, Cornell University}

\maketitle              
\begin{abstract}
Resting-state functional MRI (rsfMRI) yields functional connectomes that can serve as cognitive fingerprints of individuals.
Connectomic fingerprints have proven useful in many machine learning tasks, such as predicting subject-specific behavioral traits or task-evoked activity.
In this work, we propose a surface-based convolutional neural network (BrainSurfCNN) model to predict individual task contrasts from their resting-state fingerprints.
We introduce a reconstructive-contrastive loss that enforces subject-specificity of model outputs while minimizing predictive error.
The proposed approach significantly improves the accuracy of predicted contrasts over a well-established baseline.
Furthermore, BrainSurfCNN's prediction also surpasses test-retest benchmark in a subject identification task. \footnote{Source code is available at \url{https://github.com/ngohgia/brain-surf-cnn}}

\end{abstract}
\section{Introduction}
\begin{figure}
\includegraphics[width=\textwidth]{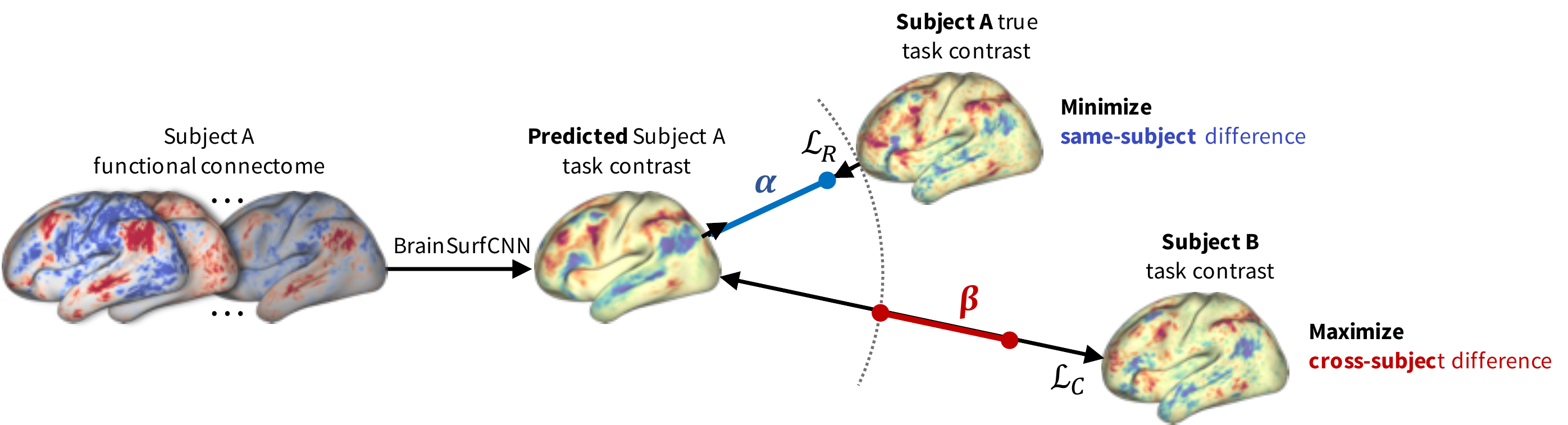}
\caption{BrainSurfCNN learns to predict an individual task contrast from their surface-based functional connectome by optimizing two objectives - minimizing the predictive error $L_{R}$
while maximizing the average difference $L_{C}$ with other subjects.
}
\label{fig:overview}
\end{figure}

Functional connectomes derived from resting-state functional MRI (rsfMRI) carry the promise of being inherent ``fingerprints'' of individual cognitive functions~\cite{biswal2010toward,kelly2012characterizing}.
Such cognitive fingerprints have been used in many machine-learning applications~\cite{khosla2019machine}, such as predicting individual developmental trajectories~\cite{dosenbach2010prediction}, behavioral traits\cite{finn2015functional}, or task-induced brain activities~\cite{tavor2016task,cole2016activity}.
In this work, we propose BrainSurfCNN, a surface-based convolutional neural network for predicting individual task fMRI (tfMRI) contrasts from their corresponding resting-state connectomes.
Figure \ref{fig:overview} gives an overview of our approach: BrainSurfCNN minimizes prediction's error with respect to the subject's true contrast map, while maximizing subject identifiability of the predicted contrast.
%
%

Prediction of individual task contrasts from rsfMRI and structural MRI features was previously explored in \cite{tavor2016task,cole2016activity} using linear regression.
In this work, we approached the same task using deep learning techniques.
This was made possible by the increased availability of rsfMRI and tfMRI imaging data from initiatives like the Human Connectome Project (HCP)~\cite{glasser2013minimal}.
Furthermore, several projects (HCP included) also repeat collection of imaging data for the same subjects on separate test and retest sessions.
Such test-retest data offer an empirical upper-bound on the reliability and replicability of neuroimaging results, including individual task contrasts.

Convolutional neural networks (CNNs) were previously used for prediction of disease status from functional connectomes~\cite{khosla2019ensemble}, albeit in volumetric space.
Instead, we used a new convolutional operator~\cite{chiyu2019spherical} suited for icosahedral meshes, which are commonly used to represent the brain cortex~\cite{van2012parcellations,fischl1999high}.
Working directly on the surface mesh circumvents resampling to volumetric space with unavoidable mapping errors~\cite{wu2018accurate}.
Graph CNN~\cite{kawahara2017brainnetcnn,niepert2016learning} is also closely related to mesh-based CNN, but there is no consensus on how pooling operates in unconstrained graphs.
In contrast, an icosahedral mesh is generated by regular subdivision of faces from a mesh of a lower resolution~\cite{baumgardner1985icosahedral}, making pooling straightforward~\cite{chiyu2019spherical}.
We also introduced a reconstructive-contrastive (R-C) loss that optimizes a dual objective of making accurate prediction while maximizing the subject's identifiability in relation to other individuals.
This objective is related to metric learning techniques~\cite{koch2015siamese,schroff2015facenet}.
Yet, to our knowledge, we are the first to examine their utility in medical image computing.

Overall, our experiments showed that the proposed BrainSurfCNN in conjunction with R-C loss yielded markedly improvement in accuracy of predicting individual task contrasts compared to an established baseline.
The proposed approach also outperforms retest contrasts in the subject identification task, suggesting that the model predictions might be useful task-evoked fingerprints for individual subjects.

\section{Materials and Methods}
\subsection{BrainSurfCNN}
\begin{figure}
\includegraphics[width=\textwidth]{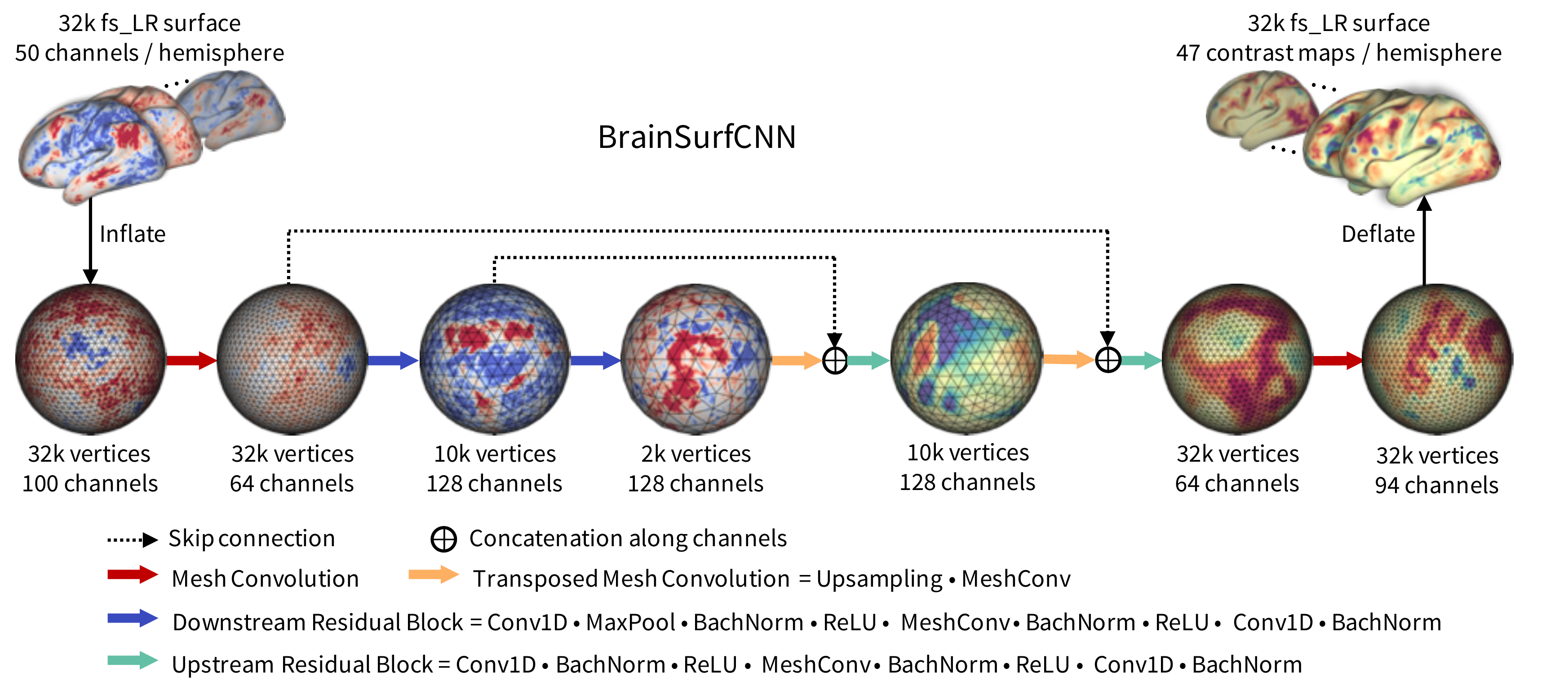}
\caption{BrainSurfCNN architecture}
\label{fig:cnn}
\end{figure}

Figure \ref{fig:cnn} shows the proposed BrainSurfCNN model for predicting task contrasts from rsfMRI-derived connectomes.
The model is based on the popular U-Net architecture~\cite{ronneberger2015u,milletari2016v} using the spherical convolutional kernel proposed in \cite{chiyu2019spherical}.
Input to the model is surface-based functional connectomes, represented as a multi-channel icosahedral mesh.
Each input channel is a functional connectivity feature, for example, the Pearson's correlation between each vertex's timeseries and the average timeseries within a target ROI.
In our experiments, the subject-specific target ROIs were derived from dual-regression of group-level independent component analysis (ICA)\cite{smith2013resting}.
The input and output surface meshes are fs\_LR meshes~\cite{van2012parcellations} with 32,492 vertices (fs\_LR 32k surface) per brain hemisphere.
The fs\_LR atlases are symmetric between the left and right hemispheres, e.g., the same vertex index in the both hemi-spheres correspond to cotra-lateral analogues.
Thus, each subject's connectomes from the two hemispheres can be concatenated, resulting in a single input icosahedral mesh with the number of channels equals twice the number of ROIs.
BrainSurfCNN's output is also a multi-channel icosahedral mesh, in which each channel corresponds to one fMRI task contrast.
This multi-task prediction setting promotes weight sharing across contrast predictions.
%
\subsection{Reconstructive-contrastive loss}

Given a mini batch of $N$ samples $B = \{\mathbf{x}_i\}$, in which $\mathbf{x}_i$ is the target multi-contrast image of subject $i$, let $\hat{\mathbf{x}}_i$ denote the corresponding predicted contrast image.
The reconstructive-contrastive loss (R-C loss) is given by:
\begin{align}
\mathcal{L}_R &= \frac{1}{N}\sum_{i=1}^{N}d(\hat{\mathbf{x}}_i, \mathbf{x}_i)
\hspace{0.5cm};\hspace{0.5cm}\mathcal{L}_C = \frac{1}{(N^2-N)/2}\sum_{\substack{\mathbf{x}_j\in{B_i}\\ j\neq{i}}}d(\hat{\mathbf{x}}_i, \mathbf{x}_j) \\
\mathcal{L}_{RC} &= \left[\mathcal{L}_R-\alpha\right]_{+} + \left[\mathcal{L}_R - \mathcal{L}_C + \beta\right]_{+}
\label{eq:loss_eq}
\end{align}
where $d(.)$ is a loss function (e.g. $l^2$-norm).
$\mathcal{L}_R$, $\alpha$ are the same-subject (reconstructive) loss and margin, respectively.
$\mathcal{L}_C$, $\beta$ are the across-subject (contrastive) loss and margin, respectively.
The combined objective enforces the same-subject error $\mathcal{L}_R$ to be within $\alpha$ margin, while encouraging the average across-subject difference $\mathcal{L}_C$ to be large such that $\left(\mathcal{L}_C - \mathcal{L}_R\right) > \beta$.

\section{Experiments}
\subsection{Data}
We used the minimally pre-processed, FIX-cleaned 3-Tesla resting-state fMRI (rsfMRI) and task fMRI (tfMRI) data from the Human Connectome Project (HCP), with the acquisition and processing pipelines described in~\cite{glasser2013minimal,smith2013resting,barch2013function}.
rsfMRI data was acquired in four 15-minute runs, with 1,200 time-points per run per subject.
HCP also released the average timeseries of independent components derived from group-level ICA for individual subjects.
We used the 50-component ICA timeseries data for computing the functional connectomes.
HCP's tfMRI data comprises of 86 contrasts from 7 task domains~\cite{barch2013function}, namely: WM (working memory), GAMBLING, MOTOR, LANGUAGE, SOCIAL RELATIONAL, and EMOTION.
Similar to \cite{tavor2016task}, redundant negative contrasts were excluded, resulting in 47 unique contrasts.

HCP released 3T imaging data of 1200 subjects, out of which 46 subjects also have retest (second visit) data.
By considering only subjects with all 4 rsfMRI runs and 47 tfMRI contrasts, our experiments included 919 subjects for training/validation and held out 39 test-retest subjects for evaluation.

\section{Baseline}
\subsection{Linear regression}
We implemented the linear regression model of \cite{tavor2016task}, given by $\mathbf{y}_{i}^{k} = \mathbf{X}^{k}_{i}\mathbf{\beta}_i^{k}$,
in which $\mathbf{y}_i^{k}$, $\mathbf{X}^{k}_{i}$, $\bm{\beta}_i^{k}$ are the vectorized activation pattern, input features, and regressor of the $k$-th parcel in the $i$-th subject.
The parcellation was derived from group-level ICA and provided by HCP.
$\mathbf{y}_i^{k}$ is a vector of length $n_k$ - the number of vertices in the $k$'th parcel in both hemispheres.
$\mathbf{X}^{k}_{i}$ is a $n_k \times M$ functional connectivity matrix, with each element computed as the Pearson's correlation between a vertex and each of the M subject-specific independent components' average timeseres (same timeseries used to compute BrainSurfCNN's input).
As in \cite{tavor2016task}, a linear regressor was fitted for every parcel and every task of each training/validation sample.
For prediction, all fitted regressors corresponding to every parcel and task contrast were averaged as one average regressor per parcel.

\subsection{Lower bound: group-average contrast}
Different tasks could exhibit different degrees of inter-individual variability and we want to assess this variability in prediction.
Thus, we computed the correlation of individual contrasts with the group average as a naive baseline.
This lower bound would be low/high for tasks with high/low inter-subject variability.

\subsection{Upper bound: retest contrast}

We used the retest (repeat) tfMRI scans to quantify the reliability of the contrast maps and assess the prediction performance of our model and the baseline. 
The retest contrasts were compared to the test (first) contrasts both in terms of overall correlation and in the subject identification task.
We consider the test-retest results as an effective upper-bound on performance.

\section{Experimental setup}
\subsection{Ensemble learning:}
\label{subsec:ensemble}
Each subject in our experiments has 4 rsfMRI runs with 1200 time-points each.
All 4800 rsfMRI time-points are often used to compute the functional connectome, resulting in one connectome per subject~\cite{tavor2016task,cole2016activity}.
On the other hand, there is evidence that stable functional connectome estimates can be computed from fewer than 1200 time-points~\cite{finn2015functional}.
We exploited this observation for data augmentation when training the models.
Specifically, each of the 4 rsfMRI runs was split into two contiguous segments of 600 time-points.
One functional connectome was computed on each segment, resulting in 8 input samples per subject.
During BrainSurfCNN training, one connectome was randomly sampled for each subject, essentially presenting a slightly different sample per subject in every batch.
For the baseline model, all 8 samples per subjects were used for training.
At test time, 8 predictions were made for each subject and then averaged for a final prediction.

\subsection{Training schedule:}
\label{subsec:training}
BrainSurfCNN was first trained for 100 epochs with a batch size of 2 with $l^2$ reconstruction loss ($L_{R}$ in Eq.\ref{eq:loss_eq}) using Adam optimizer.
Upon convergence, the average reconstructive loss $L_{R}$ and $L_{C}$ were computed from all training subjects, and used as initial values for the margins $\alpha$ and $\beta$ in Eq.\ref{eq:loss_eq} respectively.
This initialization encourages the model to not diverge from the existing reconstructive loss while improving on the contrastive loss.
We then continued training for another 100 epochs, with the same-subject margin $\alpha$ halved and across-subject margin $\beta$ doubled every 20 epochs, thus applying continuously increased pressure on the model to refine.

\subsection{Evaluation}
Pearsons' correlation coefficients were computed between the models' predicted individual task contrast maps and the tfMRI contrast maps of all subjects.
This yields a 39 by 39 correlation matrix for each contrast, where each entry is the correlation between a subject's predicted contrast (column) and an observed tfMRI contrast map (row), of same or another subject. 
The diagonal values (correlation with self) thus quantify the (within subject) predictive accuracy for a given task contrast.
The difference between diagonal and average off-diagonal values (correlation of self vs others) captures how much better one subject's prediction correlates with the corresponding subject's own tfMRI contrast compared to other subject contrasts.
From another perspective, the $i$-th subject can be identified among all test subjects by the predicted contrast if the $i$-th element of the  $i$-th row has the highest value.
For a given contrast and prediction model, we compute subject identification accuracy as the fraction of subjects with a maximum at the diagonal.

\section{Results}
\subsection{Contrasts prediction quality}
\begin{figure}
\includegraphics[width=\textwidth]{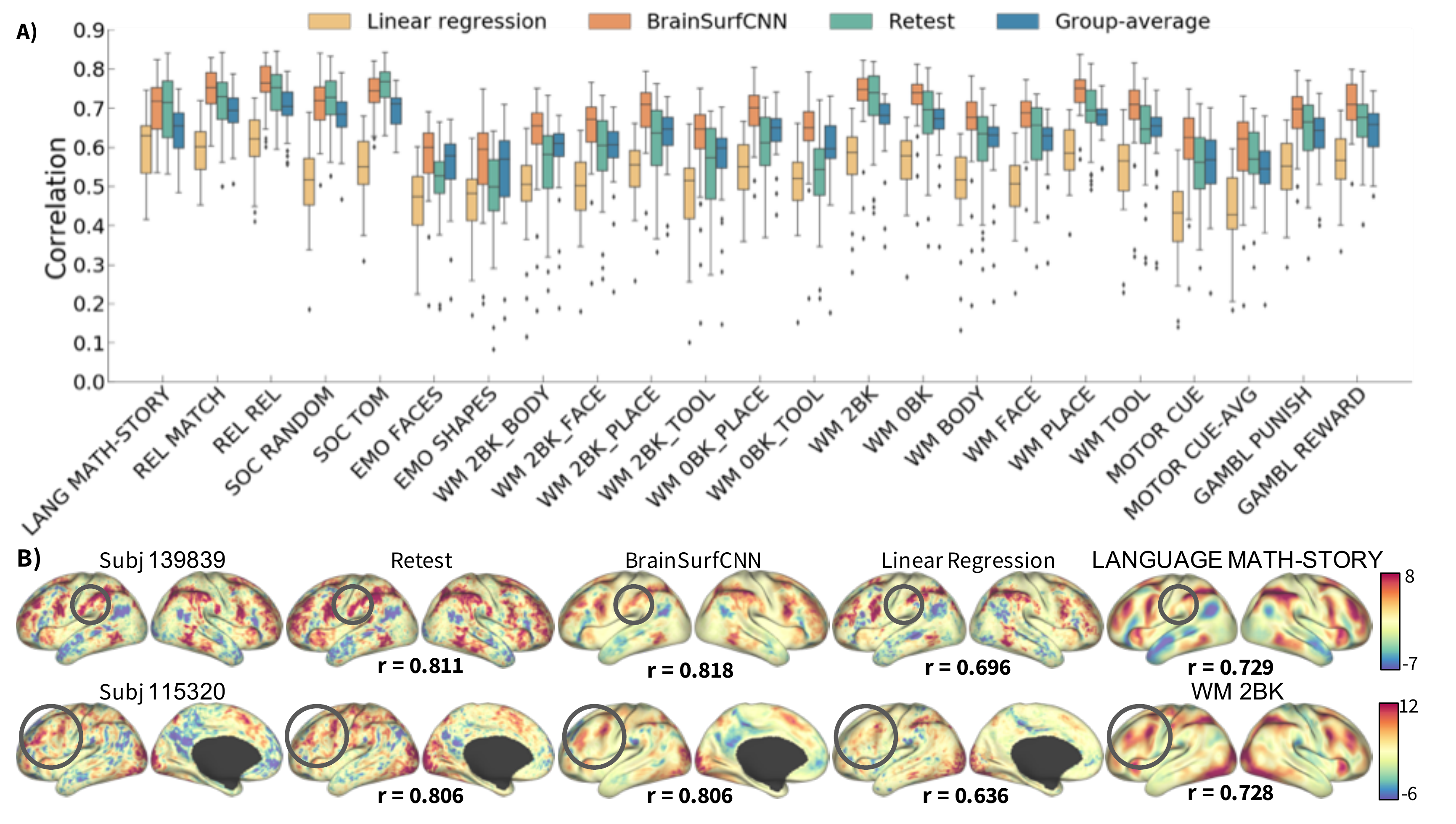}
\caption{(A) Correlation of predicted with true individual task contrasts (only reliable contrasts are shown). LANG, REL, SOC, EMO, WM, and GAMBL are short for LANGUAGE, RELATIONAL, SOCIAL, EMOTION, WORKING MEMORY and GAMBLING respectively. (B) Surface visualization for 2 task contrasts of 2 subjects. The right-most column shows the group-average contrasts for comparison.} \label{fig:subj_pred}
\end{figure}

Fig.~\ref{fig:subj_pred}A shows the correlation of models' prediction with the same subject's observed contrast maps.
Only reliably predictable task contrasts, defined as those whose average test-retest correlation across all test subjects is greater than the average across all subjects and contrasts, are shown in subsequent figures.
We include results for all contrasts in the Supplementary Materials.
Fig.~\ref{fig:subj_pred}B shows the surface visualization of 2 task contrasts for 2 subjects.
While the group-average match individual contrasts' coarse pattern, subject-specific contrasts exhibit fine details that are replicated in the retest session but washed out in the group averages (circled in Fig.~\ref{fig:subj_pred}B).
On the other hand, predictions by the linear regression model missed out the gross topology of activation specific to some contrasts (e.g. second row of Fig.~\ref{fig:subj_pred}B).
Overall, BrainSurfCNN's prediction consistently yielded the highest correlation with the individual tfMRI contrasts, approaching the upper bound of the subjects' retest reference.

\subsection{Subject identification}

\begin{figure}
\includegraphics[width=\textwidth]{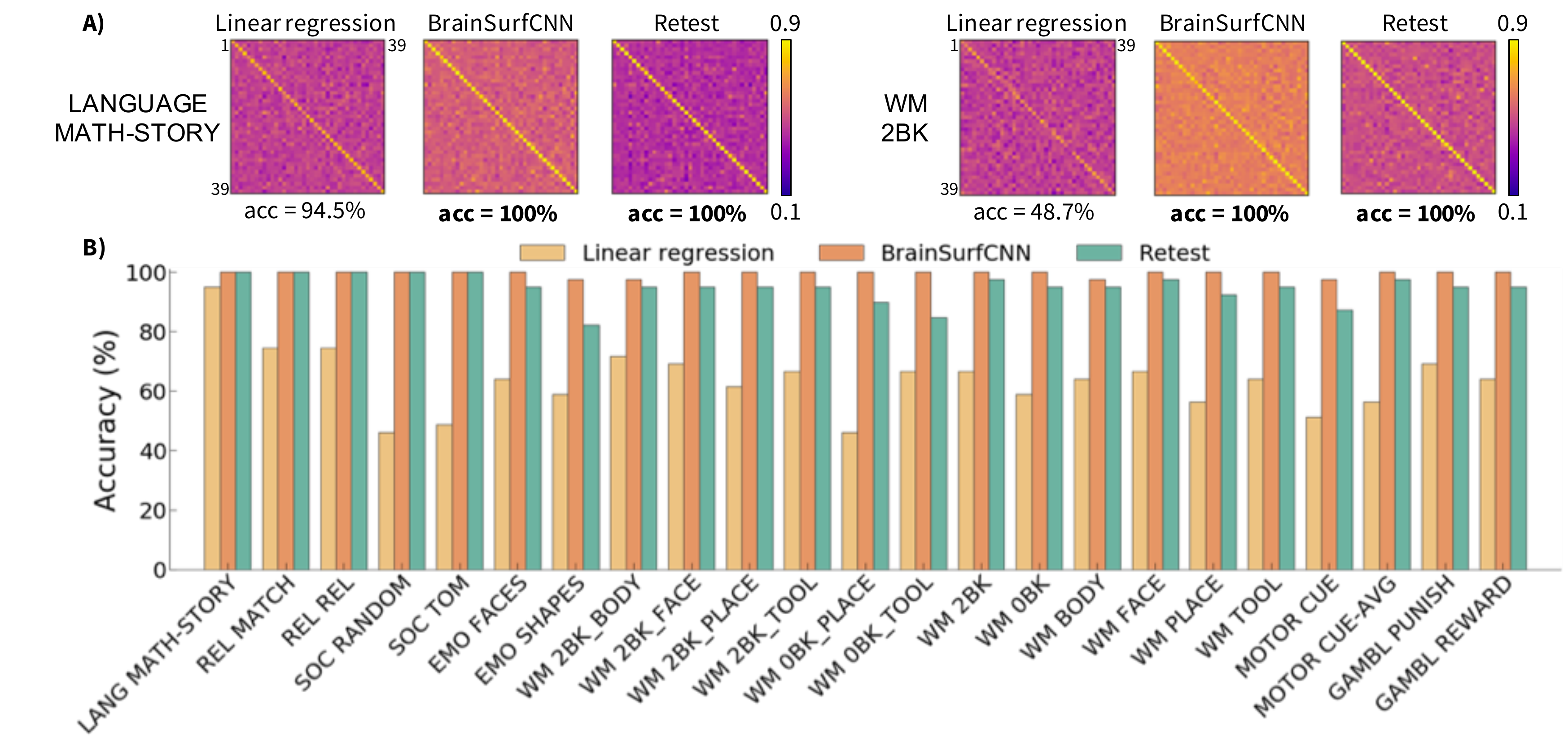}
\caption{(A) Correlation matrices (normalized) of prediction versus true subject contrasts for 2 task contrasts across 39 test subjects (B) Subject identification accuracy of predictions fobarchr 23 reliably predictable task contrasts.}
\label{fig:subj_id}
\end{figure}

Figure \ref{fig:subj_id}A shows the correlation matrices between the individual tfMRI task contrasts (rows) and the predicted task contrasts (columns) for two contrasts across all test subjects.
Similar to \cite{tavor2016task}, the matrices were normalized for visualization to account for higher variability in true versus predicted contrasts.
All matrices have dominant diagonals, indicating that the individual predictions are generally closest to same subjects' contrasts.
Across all reliable task contrasts, the task contrasts predicted by BrainSurfCNN have consistently better subject identification accuracy as compared to the linear regression model, shown in Fig.~\ref{fig:subj_id}B and the clearer diagonals in Fig.\ref{fig:subj_id}A.
%
\subsection{Ablation Analysis}
Table \ref{tabl:ablation} shows the effects of ensemble learning and reconstructive-contrastive (R-C) loss on BrainSurfCNN performance.
Ensemble learning (section~\ref{subsec:ensemble}) improves upon training with one sample per subject in predictive accuracy (diagonals of correlation matrices in Fig. \ref{fig:subj_id}), but results in smaller difference between predictions of one subject versus other subjects' contrasts (off-diagonal values).
However, the introduction of the R-C loss made the model prediction more specific to the subjects of interest.
Scheduled tuning of the loss margins (section~\ref{subsec:training}) further improved the specificity of the predictions.

\setlength{\tabcolsep}{0.5em}
\begin{table}
\centering
\caption{Effects of design choices on BrainSurfCNN predictions.}
\begin{tabular}{|l|c|c|c|}
\hline
Model                                                & \makecell{Correlation with self}  & \makecell{Correlation of \\self minus other } \\\hline 
1 sample/subject                                     & $0.64\pm{0.11}$                   & $0.060\pm{0.036}$                                    \\\hline 
Ensemble (8 samples/subject)                         & $0.66\pm{0.11}$                   & $0.046\pm{0.026}$                                   \\\hline 
Ensemble + 100 more epochs with $l^2$ loss           & $0.66\pm{0.11}$                   & $0.048\pm{0.026}$                                    \\\hline 
Ensemble + R-C Loss                                  & $0.66\pm{0.11}$                   & $0.081\pm{0.045}$                                   \\\hline 
Ensemble + scheduled R-C Loss                        & $0.66\pm{0.11}$                   & $0.087\pm{0.047}$                                    \\\hline 
Retest                                               & $0.61\pm{0.13}$                   & $0.181\pm{0.089}$                                    \\\hline 
\end{tabular}
\label{tabl:ablation}
\end{table}

\section{Discussion and Conclusion}
Cognitive fingerprints derived from rsfMRI have been of great research interest~\cite{finn2015functional,amico2018quest}.
The focus of tfMRI, on the other hand, has been mostly on seeking consensus of task contrasts across individuals.
Recent work exploring individuality in task fMRI mostly utilized sparse activation coordinates reported in the literature~\cite{yeo2015functional,ngo2019beyond} and/or simple modeling methods~\cite{tavor2016task,cole2016activity,amico2018quest}.
In this paper, we presented a novel approach for individualized prediction of task contrasts from functional connectomes using surface-based CNN.
In our experiments, the previously published baseline model~\cite{tavor2016task} achieved lower correlation values than the group-averages, which might be due to the ROI-level modeling that misses relevant signal from the rest of the brain.
The proposed BrainSurfCNN yielded predictions that were overall highly correlated with and highly specific to the individuals' tfMRI constrasts .
We also introduced a reconstructive-contrastive (R-C) loss that significantly improved subject identifiability, which are on par with the test-retest upper bound.

We are pursuing several extensions of the current approach.
Firstly, we plan to extend the predictions to the sub-cortical and cerebellar components of the brain.
Secondly, BrainSurfCNN and R-C loss can be applied to other predictive domains where subject specificity is important, such as in individualized disease trajectories.
Lastly, we can integrate BranSurfCNN's prediction into quality control tools for tfMRI when retest data are unavailable.

Our experiments suggest that a surface-based neural network can effectively learn useful multi-scale features from functional connectomes to predict tfMRI contrasts that are highly specific to the individual.

\section{Acknowledgements}
This work was supported by NIH grants R01LM012719 (MS), R01AG053949 (MS), R21NS10463401 (AK), R01NS10264601A1 (AK), the NSF NeuroNex grant 1707312 (MS), the NSF CAREER 1748377 grant (MS), Jacobs Scholar Fellowship (GN), and Anna-Maria and Stephen Kellen Foundation Junior Faculty Fellowship (AK).
The authors would like to thank the reviewers for their helpful comments, Ms. Hao-Ting Wang for her pointers on preprocessing HCP data and Mr. Minh Nguyen for his comments on the early drafts.

\bibliographystyle{unsrt}
\bibliography{main}

\appendix
\newpage
\pagenumbering{arabic}
\setcounter{page}{1}
\renewcommand{\thefigure}{S\arabic{figure}}
\setcounter{figure}{0}
\begin{center}
\large\textbf{Supplementary Materials for ``From Connectomic to Task-evoked Fingerprints: Individualized Prediction of Task Contrasts from Resting-state Functional Connectivity''}
\end{center}
\section{Contrast prediction quality for all task contrasts}
\begin{figure}
\includegraphics[width=\textwidth]{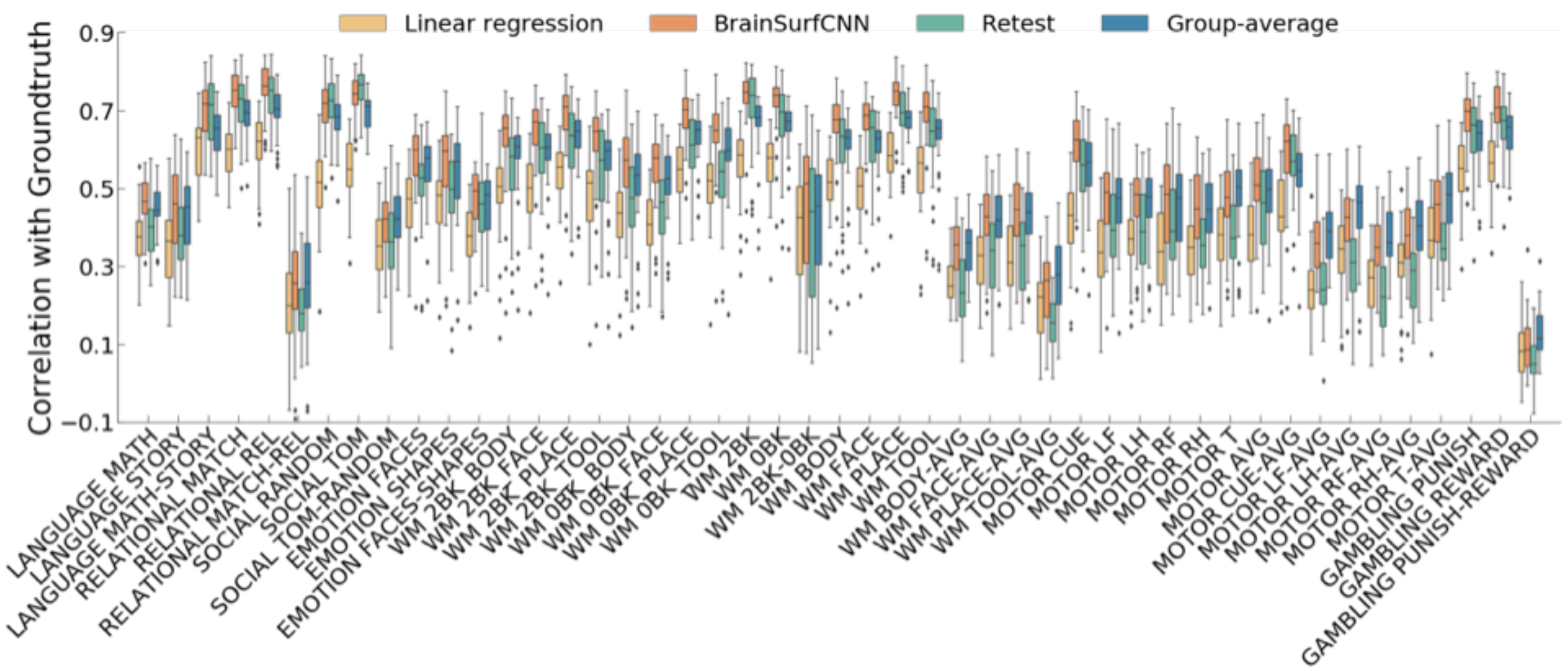}
\caption{Correlation of predicted with true individual task contrasts.}\label{fig:appendix_subj_pred}
\end{figure}
\section{Subject identification accuracy of model predictions}
\begin{figure}
\includegraphics[width=\textwidth]{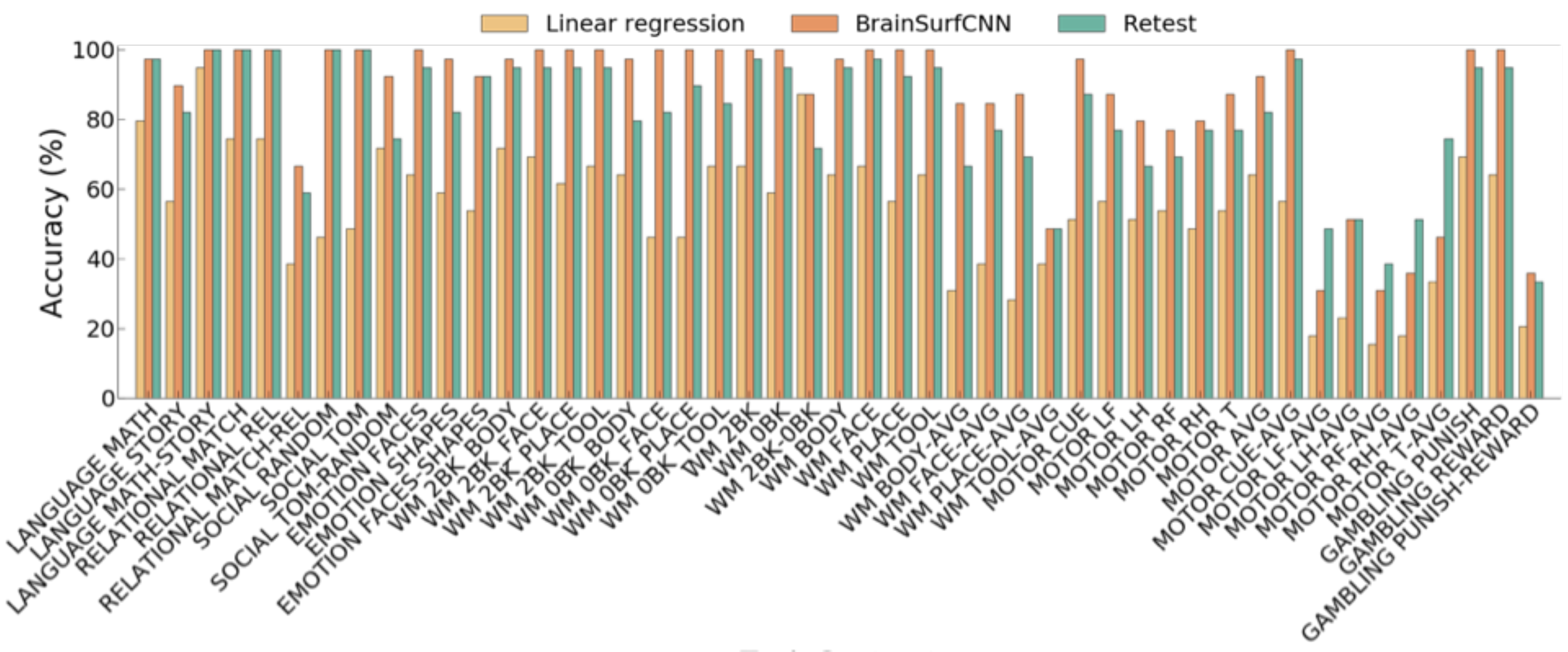}
\caption{Subject identification accuracy of model predictions.}\label{fig:appendix_subj_id}
\end{figure}
\end{document}